\documentclass[letterpaper, 10 pt, conference]{ieeeconf}  % Comment this line out if you need a4paper

\IEEEoverridecommandlockouts                              % This command is only needed if 
\makeatletter
\def\endthebibliography{%
  \def\@noitemerr{\@latex@warning{Empty `thebibliography' environment}}%
  \endlist
}

\usepackage{graphics} % for pdf, bitmapped graphics files
\usepackage{epsfig} % for postscript graphics files
\usepackage{subfigure}
\usepackage{amsmath} % assumes amsmath package installed
\usepackage{epstopdf}
\usepackage{hyperref}
\usepackage{lipsum,amsmath,multicol}
\usepackage{float}
\usepackage{algorithm}
\usepackage{algorithmicx}
\usepackage{algpseudocode}
\usepackage{hyperref}

\title{\LARGE \bf
Model Predictive Control for Autonomous Driving considering Actuator Dynamics
}

\author{Mithun Babu$^{1}$, Raghu Ram Theerthala$^{1}$, Arun Kumar Singh$^{2}$, Baladhurgesh B.P.$^{1}$,\\  Bharath Gopalakrishnan$^{1}$, K. Madhava Krishna $^{1}$% <-this % stops a space
\thanks{$^{1}$ Robotics Research Center, International Institute of Information Technology, Hyderabad, India
      }%
\thanks{$^{2}$ Department of Hydraulics and Automation, Tampere University of technology, Finland
      }%
}

\begin{document}

\maketitle
\thispagestyle{empty}
\pagestyle{empty}

%%%%%%%%%%%%%%%%%%%%%%%%%%%%%%%%%%%%%%%%%%%%%%%%%%%%%%%%%%%%%%%%%%%%%%%%%%%%%%%%
\begin{abstract}

In this paper, we propose a new model predictive control (MPC) formulation for autonomous driving. The novelty of our MPC stems from the following results. Firstly, we adopt an alternating minimization approach wherein linear velocities and angular accelerations are alternately optimized. We show that in contrast to the joint optimization, the alternating minimization exploits the structure of the problem better, which in turn translates to reduction in computation time. Secondly, our MPC explicitly incorporates the time dependent non-linear actuator dynamics that captures the transient response of the vehicle for a given commanded velocity. This added complexity improves the predictive component of MPC resulting in improved margin of inter-vehicle distance during maneuvers like overtaking, lane-change, etc. Although, past works have also incorporated actuator dynamics within MPC, there has been very few attempts towards coupling actuator dynamics to collision avoidance constraints through the non-holonomic motion model of the vehicle and analyzing the resulting behavior. We use a high fidelity simulator to benchmark our actuator dynamics augmented MPC with other related approaches in terms of metrics like inter-vehicle distance, trajectory smoothness, and velocity overshoot.
\end{abstract}

\section{Introduction}

Autonomous driving represents an interesting research problem which is at the intersection of many different fields including robotics and control. Motion planning is a core component of any autonomous driving set-up. Off late, there has been an upsurge in applying model predictive control (MPC) based formulation for navigation of autonomous vehicles \cite{mpc_1}, \cite{mpc_2}, \cite{mpc_3}. The motivation behind this is clear. MPC provides a unified optimization based approach which can handle arbitrary constraints on state and control while minimizing a user-defined cost function. In spite of the recent success of MPC on autonomous driving, some key bottlenecks still remain. Among these include improving the computational efficiency of the underlying optimization and improving the predictive component of MPC to better reflect the actual behavior of the vehicle. In this paper, we make contributions towards both these problems.

\begin{figure}[!tbh]
  \centering
   \subfigure[]{
    \includegraphics[width= 3in] {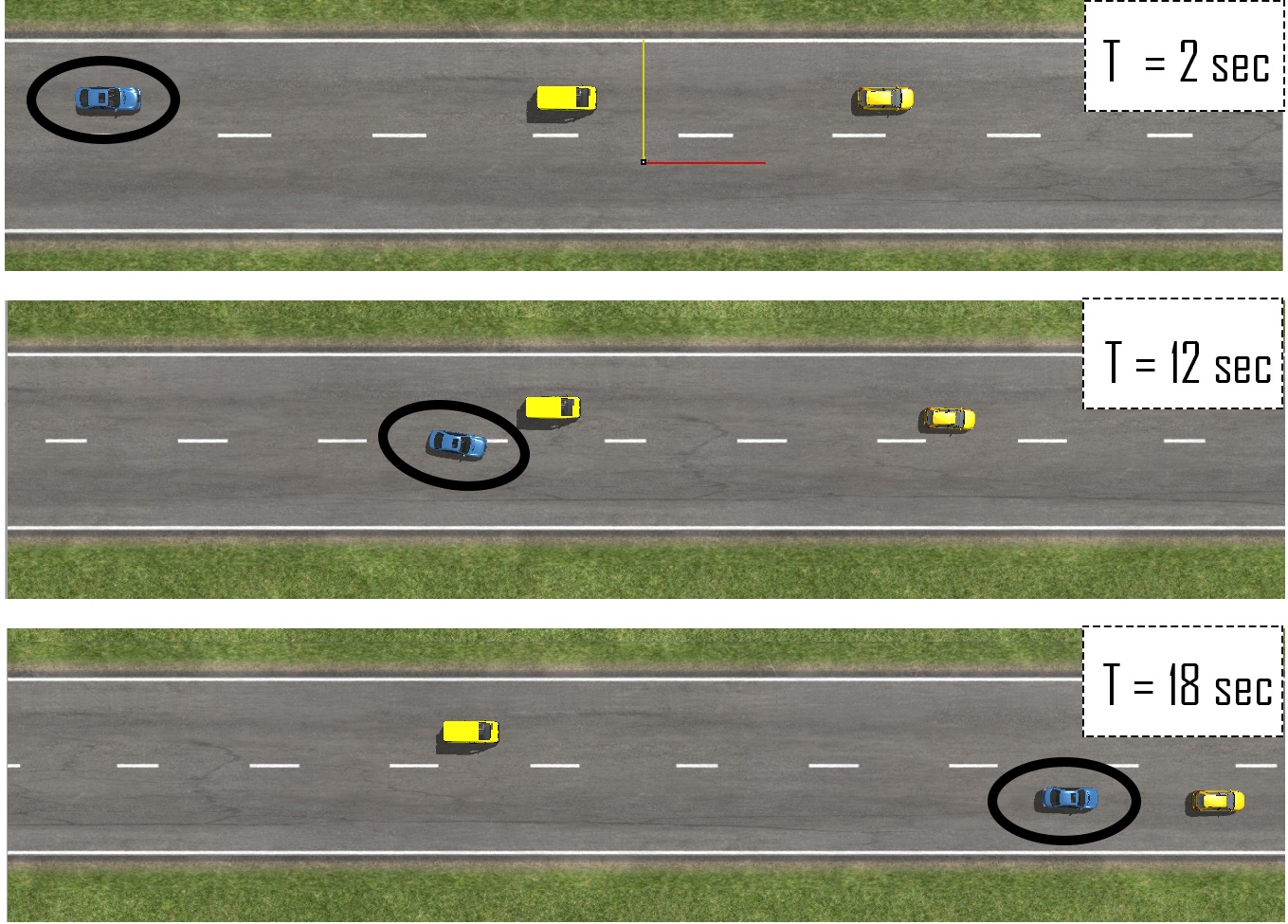}
    \label{ex_plot1}
   }

   \caption{Ego vehicle (in blue) performing lane change maneuver based on the MPC framework proposed in the paper. Simulations of several such scenarios is available at \url{https://researchweb.iiit.ac.in/~mithun.babu/acc_2019.mp4}}          
\end{figure}

On the optimization front, we make a case for adopting the alternating minimization (AM) approach which alternately searches in the space of angular accelerations and linear velocities. More precisely, the optimization operates in two separate layers wherein at the first layer, the angular accelerations are optimized while fixing the linear velocities. Subsequently, at the second layer, the linear velocities are optimized while the angular accelerations are fixed to the values obtained at the first layer and so on. We show that AM approach has a very distinct advantage. For a given angular acceleration profile, the non-holonomic motion model reduces to an affine form with respect to linear velocity. This in turn leads to a difference of convex structure in the velocity optimization layer which allows us to solve it efficiently using the  \emph{convex-concave procedure} \cite{boyd_mcp}, \cite{boyd_sqp}. Although AM has been extensively used in robotics and machine learning applications, we are not aware of any existing works which applies it to simplify trajectory optimization with non-holonomic motion model.

%
% The actuator dynamics is itself estimated from the data collected on the CARSIM simulator \cite{carsim_na}.

Our second contribution lies in incorporating actuator dynamics in the underlying optimization of our MPC. To be precise, we incorporate the time dependent non-linear mapping between the commanded velocity and the actual body velocity attained by the vehicle. The idea of MPC with actuator dynamics is not new. However, in the context of autonomous vehicles, existing works around this idea have been mostly restricted to  either only path tracking control \cite{mpc_actuator1} or have considered collision avoidance only along the longitudinal direction of motion \cite{nikolce_mpc_actuator}. In contrast, the novelty of our approach stems from how we relate the actuator dynamics to the collision avoidance constraints through the non-holonomic motion model.

On the implementation side, we show the usefulness of our AM approach by showing reduction in computation time over the joint formulation that simultaneously optimizes linear velocities and angular accelerations. Furthermore, we use a high fidelity simulator to compare our actuator dynamics augmented MPC with a more conventional formulation with no-actuator dynamics and piece-wise constant acceleration input. The specific metrics used for comparison include margin of inter-vehicle distance and velocity oscillation during standard maneuvers like lane-change, overtaking and lane following.
The rest of the paper is organized as follows: Section \ref{relatedworks} gives an overview of the current work in this field, Section \ref{traj_opt} describes the actuator dynamics in detail along with the formulation of cost function and constraints. Section \ref{AM}, describes the alternating minimization routine and section \ref{results_disc} gives a detail evaluation of our AM approach.

\section{Related Works}
\label{relatedworks}
In this section, we present a review of MPC based autonomous driving, especially focusing on the underlying optimization of the MPC and whether or not it considers actuator dynamics.

MPC for autonomous driving is an active area of research and there exists a lot of literature that justifies its potential. See \cite{mpc_3}, \cite{ad_survey}, \cite{ad_survey2} for some excellent surveys. For better comparison with our proposed work, we classify the existing MPC formulations into two categories. Those in the first category like \cite{mpc_1}, \cite{nonlinear2}  formulate the underlying optimization of the MPC as a rigorous non-linear programming problem and then use iterative techniques like sequential convex programming \cite{boyd} for the solution. The advantage of these approaches is that they consider the exact motion model of the vehicle and thus, are guaranteed to produce a kinematically feasible trajectory. The so called warm-start or using trajectories obtained at the past iteration to initialize the optimization at the current iteration has been shown to speed up computation in practice. The formulations in the second category directly works with an affine approximation of the vehicle motion model \cite{linear1}, \cite{linear2}. Consequently, the optimization becomes simpler although at the expense of obtaining trajectories that may not be kinematically feasible or even collision free. Our AM based approach falls into the first category. But in contrast to existing works, we try to exploit the inherent structure in the optimization leading to tangible computational improvements. 

All the above cited works employ a sophisticated trajectory optimization but do not consider the actuator dynamics within the optimization.  Most of the current MPC formulations with  actuator dynamics either have not considered collision avoidance or have worked with simplified collision benchmarks. For example, \cite{mpc_actuator1} claims improved tracking performance by incorporating actuator dynamics within the MPC. Authors in  \cite{nikolce_mpc_actuator} formulate cooperative adaptive cruise control for a fleet of vehicles and have considered collision avoidance but only along the longitudinal direction of motion. Along the same line, \cite{luo_md} builds a MPC  with actuator dynamics but does not consider collision avoidance.

%To the best of the author's knowledge this work is the first of its kind that brings in an alternating Minimization procedure that seamlessly integrates itself with the actuator dynamics thus claiming distinct advantages.

\section{Problem Formulation}\label{traj_opt}

In this section, we formulate the underlying optimization associated with our MPC formulation. We begin by introducing the motion model and actuator dynamics followed by systematic description of cost function and constraints of our optimization problem.

\subsection{Motion model}\label{motion_model_theory}
\noindent The discrete time model of a  non-holonomic autonomous vehicle  is given by the following set of equations.

\begin{equation}
\begin{array}{ccl}
    x(t_{i+1}) &=& x(t_i) + v(t_i)\cos\theta(t_i)\Delta t.\\
    y(t_{i+1}) &=& y(t_i) + v(t_i)\sin\theta(t_i)\Delta t.\\
    \theta(t_{i+1}) &=& \theta(t_i)+\dot{\theta}(t_i)\Delta t+ \frac{1}{2} \ddot{\theta}(t_i)\Delta t^2\\
    \dot{\theta}(t_{i+1}) &=& \dot{\theta}(t_i)+\ddot{\theta}(t_i)\Delta t\\
    \ddot{\theta}(t_i) &=& u_1(t_i)\\
    v(t_i) &= & f_{act}(t,v_c(t_i), v(t_{i-1}))\\
    v_c(t_i)& = & u_2(t_i)
\end{array}
      \label{nonhol_motion_model}
\end{equation}

\noindent Where, $x(t_i)$, $y(t_i)$ and $\theta(t_i)$ are respectively the position and heading of the vehicle. The term $v(t_i)$ represents the linear velocity of the vehicle while $\dot{\theta}(t_i)$ corresponds to  the angular velocity at time  $t_i$.  As shown the motion model is controlled by angular accelerations, $\ddot{\theta}(t_i)$ and velocity commands, $v_c(t_i)$. Furthermore, we assume that commanded velocity gets translated to actual body velocity ($v(t_i)$)of the vehicle through a non-linear time dependent function, $f_{act}(.)$.

\subsection{Actuator Dynamics}\label{act_dyn_theory}
\noindent In the context of the proposed work, the role of actuator dynamics, $f_{act}(.)$ is to model how actual body velocity varies with respect to the commanded velocity over time. As a simple example, consider a $f_{act}(.)$ which relates $v(t)$ and $v_c(t_i)$ linearly over time. Thus, the relationship between $v(t)$ and $v_c(t_i)$ takes the following form:

\begin{equation}
v(t) = v(t_{i-1})+\overbrace{\frac{v_c(t_i)-v(t_{i-1})}{t_i-t_{i-1}}}^{a(t_{i-1})}(t-t_{i-1}), \forall t \in (t_{i-1} \hspace{0.2cm} t_{i}]
\label{linear_act}
\end{equation} 

\noindent It is clear that (\ref{linear_act}) is equivalent to non-holonomic motion model with piece-wise constant acceleration $a(t_{i-1})$ as the control input. In our work, we adopt a more expressive $f_{act}(.)$ which accounts for the inherent relationship between $v(t_i)$ and $v_c(t_i)$.

\begin{figure}[!tbh]
  \centering
    \subfigure[]{
    \includegraphics[width= 8.5cm, height=3cm] {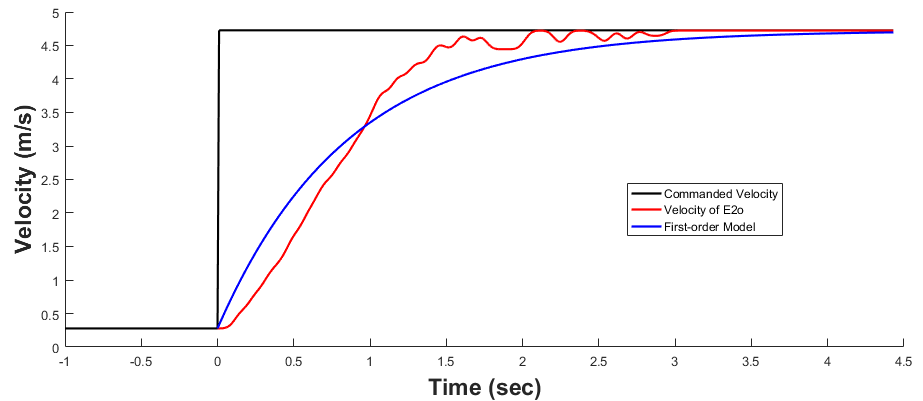}
    \label{opt_p1}
    }  
   \subfigure[]{
    \includegraphics[width= 8.5cm, height=3cm] {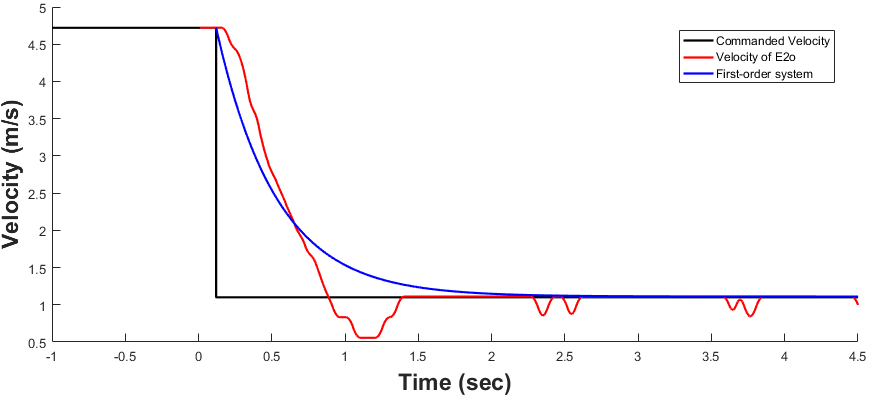}
    \label{opt_p2}
    }
 \subfigure[]{
    \includegraphics[width= 8.5cm, height=3cm] {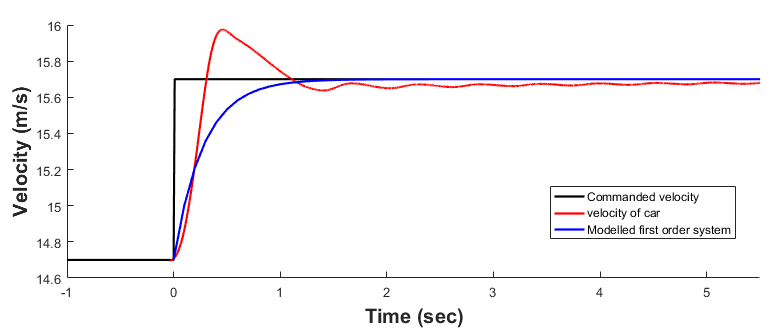}
    \label{opt_p3}
   }   
     \caption{Figures show response to a step velocity input. The commanded velocity is shown in black, the response in red and the fitted first order model in blue. (a) and (b) show results from our autonomous vehicle hardware a Mahindra E20 \cite{rrc_ad}. Predominantly, the observed response was similar to a first order system. However, occasionally during acceleration, the vehicle did show velocity overshoots. The response from our simulator shown in (c) is more typical of a second order system.}   
     \label{exp_p1}         
\end{figure}

\noindent \textbf{First order system:} We conducted several experiments with  our autonomous vehicle hardware \cite{rrc_ad} and a high-fidelity simulator CARSIM \cite{carsim_na} to analyze the response to a step velocity input. Some sample results are presented in Fig. \ref{opt_p1}-\ref{opt_p3}. Predominantly, our autonomous vehicle shows actuator dynamics similar to a first order system (\ref{opt_p1}). However, during de-accelerations (Fig. \ref{opt_p2}), we  observed occasional velocity overshoots indicating that the actual dynamics could have multiple modes. The response from our simulator (\ref{opt_p3}) is more typical of a second order system.  

Motivated by the results from our autonomous vehicle hardware, we model our actuator dynamics as a first order system (\ref{first_order_model}), where $v(t_{i-1})$ is the velocity acquired by the system at $t_{i-1}$ instant and $v(t)$ is time varying response for a given step input. Model parameter $\tau$ is called time constant of the system. Finding $\tau$ analytically is difficult as many components of our vehicle actuation mechanism are difficult to model. So, we adopted a data-driven approach and estimated $\tau$ following a linear regression based on the response to different velocity step inputs. We note that our first order actuator dynamics is a simplified model and it leads to a simpler optimization problem (see equation (\ref{vel_act_model})). Moreover, our extensive simulations show that even with a simulator which predominantly has second order actuator dynamics, our first order model could ensure collision avoidance with high fidelity. Finally, a first order model has only one parameter and thus require simpler system identification as compared to a second order actuator dynamics.

\small
\begin{equation}
\begin{array}{lr}
 v(t) = v_c(t_i) + \bigg(v(t_{i-1}) - v_c(t_i)\bigg)\exp^{\frac{-(t-t_{i-1})}{\tau}} t \in (t_{i-1},t_{i}] 
\end{array},
\label{first_order_model}
\end{equation}
\normalsize

\noindent \textbf{Expression for vehicle velocity:} If the system is subjected to a set of $n$ commanded velocity inputs, then vehicle velocity  $v(t)$  after $n$ time-steps each of $\Delta t$ duration, can be represented as

\small
{\begin{equation}
\begin{array}{lr}
 v(t_n) = (\sum_{i=1}^{n}v_{c}(t_i)(1-m_i)\prod_{l=i+1}^{n}m_l)+v_o\prod_{l=0}^{n}m_l.  
\end{array}
\label{vel_act_model}
\end{equation}}
\vspace{-0.3cm}
\begin{equation}
\begin{array}{lr}
m_i=\exp^{-\Delta t/\tau}
\end{array}
\label{M_i}
\end{equation} 
\normalsize
The usefulness of (\ref{vel_act_model}) stems from the fact that it provides time varying vehicle velocity response purely as a function of the commanded velocity inputs $v_c(t_i)$. Importantly, (\ref{vel_act_model}) is affine with respect to $v_c(t_i)$.

\subsection{Optimization}\label{Cost_func_constr_theory}
\noindent The underlying optimization of our MPC  can be described in the following manner:

\begin{subequations}
\begin{align}
\arg\min_{\ddot{\theta}(t_i), v_c(t_i)} J = J_{smooth}+J_{goal}\label{cost}\\
\textbf{x}(t_{i+1}) = \textbf{f}(\textbf{x}(t_i), \textbf{u}(t_i)). \label{motion_mod}\\
\dot{\theta}_{min} \leq \dot{\theta}(t_i)\leq \dot{\theta}_{max} \label{thetadotmax}\\
\ddot{\theta}_{min}\leq \ddot{\theta}(t_i)\leq \ddot{\theta}_{max}\label{angularaccmax}\\
v_c(t_i)\leq v_{max} \label{velmax}\\
a_{min} \leq \frac{v_c(t_{i})-v(t_{i-1})}{\Delta t} \leq a_{max} \label{accelmax}\\
-\kappa_{max}v(t_i)\leq \dot{\theta}(t_i)\leq \kappa_{max}v(t_i)\label{curvmax}\\
c_{obst}(x(t_i), y(t_i), x_i(t_i), y_i(t_i), R_i)\leq 0\label{collavoid}
\end{align}
\end{subequations}

\small
\begin{equation}
J_{smooth} = \overbrace{\sum_{i=1}^{N} \ddot{\theta}(t_i)^2}^{J_{\theta}}+\overbrace{\sum_{i=1}^{N}(\frac{v_c(t_{i-1})-2 v_c(t_{i})+v_c(t_{i+1})}{{\Delta t}^2})^2}^{J_v}
\label{cost_smooth}
\end{equation}
\normalsize

\begin{equation}
J_{goal} = (x(t_N)-x_f)^2+(y(t_N)-y_f)^2 + (\theta(t_N)-\theta_f)^2
\label{cost_terminal}
\end{equation}

\noindent Where, $\textbf{x}(t_i)= [x(t_i), y(t_i), \theta(t_i), \dot{\theta}(t_i)]$ and represents the state of the system. The vector valued function $\textbf{f}(.)$ is just a compact representation of (\ref{nonhol_motion_model}), where we have used  $\textbf{u}(t_i)= [u_1(t_i),u_2(t_i)]$ as mentioned in \ref{nonhol_motion_model}. The cost function (\ref{cost}) is a summation of  smoothness and goal reaching  costs (\ref{cost_smooth})-(\ref{cost_terminal}). As can be seen from (\ref{cost_smooth}), smoothness cost penalizes high value of angular accelerations and jerk modeled as a second order finite difference of linear velocity. The terminal cost ensures that the obtained trajectory terminates as close as possible to the goal position $(x_f, y_f)$. The equality (\ref{motion_mod}) constrains the control variables and states to be compatible with the motion model of the robot. The inequalities (\ref{thetadotmax})-(\ref{angularaccmax}) represent the bounds on angular velocities and accelerations respectively. Inequality (\ref{velmax}) ensures that the commanded velocity is less than the  physical limit of the vehicle. In (\ref{accelmax}) we have modeled acceleration as a finite difference between the vehicle velocity at $t_{i-1}$ and the next commanded velocity. Inequalities, (\ref{curvmax}) represent the curvature bounds for the vehicles. Note how these have been written with respect to the actual body velocity and not the commanded velocity. Inequalities (\ref{collavoid}) models the collision avoidance constraints and has the following algebraic form:

%Following \cite{mpc_1},\cite{viswa} the same form can be leveraged for polygonal shapes(car like) by approximating them through multiple overlapping circles.

\begin{equation}
c_{obst}(.)\leq 0: -(x(t_i)-x_i(t_i))^2-(y(t_i)-y_i(t_i))^2+R_i^2\leq 0.
\label{coll_avoid_form}
\end{equation}

\noindent Where, $x_i(t_i)$ and $y_i(t_i)$ describe the position of $i^{th}$ obstacle. Following \cite{mpc_1}, \cite{viswa}, we model our ego-vehicle and the neighboring obstacles as an overlap of circles. Consequently, $R_i$ represents the combined radius of each circle of the  ego-vehicle and $i^{th}$ obstacle. For static obstacles, the position would be independent of $t_i$. The form of (\ref{coll_avoid_form}) assumes that the vehicle and the obstacles are both modeled as circles.  It is convenient to obtain an affine approximation for (\ref{coll_avoid_form}) by linearizing it around a guess trajectory $(\hat{x}(t_i), \hat{y}(t_i))$. In (\ref{affine_approx}), $\bigtriangledown_x$,  $\bigtriangledown_y$ stands for partial derivative of $c_{obst}(.)$ with respect to $x(t_i)$ and $y(t_i)$ respectively. Further, $\hat{c}(.)$ is obtained by evaluating right hand side of (\ref{coll_avoid_form}) at $(\hat{x}(t_i), \hat{y}(t_i))$.

\begin{equation}
\resizebox{1\hsize}{!}{$^{affine}c_{obst}(.)= \hat{c}_{obst}+\bigtriangledown_x(x(t_i)-\hat{x}(t_i))+\bigtriangledown_y(y(t_i)-\hat{y}(t_i))$}
\label{affine_approx}
\end{equation}
\normalsize

\noindent The core complexity of optimization (\ref{cost})-(\ref{collavoid}) stems from the non-linear  motion model (\ref{motion_mod}). This is because, for an affine motion model, (\ref{affine_approx}) turns out to be globally valid convex approximation of the original collision avoidance constraints (\ref{coll_avoid_form}). \cite{boyd_sqp}, \cite{boyd_mcp}. In other words, satisfaction of (\ref{affine_approx}) ensures satisfaction of (\ref{coll_avoid_form}) as well. However, the non-linearity of non-holonomic motion model destroys this structure and makes the optimization more difficult.

In the next section, we propose an optimization routine which tries to explore as much as possible the inherent structure of the problem. The core idea hinges on the observation that if the heading trajectory of the vehicle is fixed, the motion model becomes affine with respect to linear velocities.

%These  guess commanded velocities are used to update $\tau $ (line 18) using the function fitted in \ref{carsim_data}. This in turn updates the relation between $v_c(t)$ and $v(t)$ through motion model in next iteration. Convergence of similar algorithms with LTV routines is described in \cite{pintro}.

\section{Alternating Optimization} 
\label{AM}

Algorithm \ref{algo1} summarizes the proposed alternating optimization routine. It starts with (line 1) choosing an initial guess for commanded velocity $\hat{v}_c^k(t)$, angular acceleration $\hat{\ddot{\theta}}^k(t)$ and a counter $k$ along with two positive weights $w_{\theta}, w_{v}$. The function $InitialTraj(.)$ (line 3) computes an initial guess for position and heading trajectory, $(\hat{x}^k(t), \hat{y}^k(t), \hat{\theta}^k(t))$ using $\hat{v}_c^{k-1}(t)$ and $\hat{\ddot{\theta}}^{k-1}(t)$ in the motion model (\ref{nonhol_motion_model}). Lines 6 and 12 represent the angular acceleration and linear velocity optimization layer respectively. Both the layers continue till the change in the cost functions between subsequent iteration is greater than the threshold $\epsilon$ and the collision avoidance constraints are not satisfied. The optimal solution obtained after each layer is used to update the initial guesses of angular accelerations and commanded velocities (lines 11 and 17).

\subsection{Angular Acceleration Layer}
\noindent The angular acceleration layer is obtained by extracting the $\theta(t)$ dependent terms from the optimization (\ref{cost})-(\ref{collavoid}). The following points are worth pointing out.

\begin{itemize}
\item First, note the motion model $f^{\theta}(.)$ which is obtained by first order Taylor series expansion of the first two equations in (\ref{nonhol_motion_model}) around $\hat{\theta}^k(t)$. Consequently, we obtain a motion model which is affine with respect to heading angle $\theta(t)$.

\item  The affine approximation  holds only  in the vicinity of $\hat{\theta}^k(t)$. Thus, a trust region needs to be incorporated to ensure that $\hat{\theta}^k(t)$ and $\hat{\theta}^{k+1}(t)$ are sufficiently close to each other. The last inequality in line 6 of Algorithm \ref{algo1} which puts a box constraints on $\theta(t)$  serves this purpose. The trust region is modified and guess point is updated at each iteration as discussed in \cite{boyd_sqp}.

\item The collision avoidance constraints have been augmented with a non-negative slack variable $s_{\theta}(t)$. This is to ensure that the Algorithm \ref{algo1} continues to make progress towards the optimal solution even if the initial guess trajectory $(\hat{x}^k(t), \hat{y}^k(t))$ renders $^{affine}c(.)$ infeasible (or in other words is not collision free). Consequently, we also incorporate a penalty on the slack variables in the cost function. The weights $w_{\theta}$ of the penalty is sequentially increased using a positive factor $\delta$ till $c_{obst}(.)>0$. 
\end{itemize}

\subsection{Linear Velocity Layer}
\noindent This layer has only such terms from the cost and constraint functions of optimization, (\ref{cost})-(\ref{collavoid}) which explicitly depends on the $v_c(t_i)$. The following key points should be noted.

\begin{itemize}
\item Note, the motion model, $\textbf{f}^{v}(.)$ which has been obtained from (\ref{nonhol_motion_model}) by using $\hat{\theta}^k(t), \hat{\dot{\theta}}^k(t), \hat{\ddot{\theta}}^k(t)$ obtained by the previous angular acceleration layer. 

%We linearize the model around $\hat{v}^k(t)$ using first-order Taylor series expansion similar to what has been done previously in angular velocity layer.
%
%\item A trust region constraint is added to make this approximation valid and the linerization point is updated every iteration so that our approximated model stays close to original model. 

\item The collision avoidance constraints are augmented with non-negative slacks $s_{v}(t)$ similar to angular acceleration layer. The penalty on the slacks also follows the same reasoning.

\item Note that this layer does not have any trust region constraints. This is because in this layer we do not make any linearization which in turn is due to the motion model being already affine in terms of forward velocity. The physical implication of this is that the velocity optimization layer can take as large as possible step towards the optimal solution \cite{boyd_sqp}. In contrast, the progress of the angular acceleration layer is limited by the size of the trust region.

\end{itemize}

\subsection{MPC}
\label{init_traj}

\noindent Algorithm \ref{algo1} is solved in a receding horizon manner to serve as our MPC. At the start of the MPC, the Algorithm \ref{algo1} is intialized with the output of a high-level planner such as \cite{sharmas}. In the subsequent iteration of the MPC, we adopt the warm-start approach where we initialize with the trajectory and controls obtained in the previous iteration.

%Once we have executed first $m$ control outputs provided by the optimization routine we observe the current state, obstacle information through on-board sensors and use this output as an initial estimate for our next iteration. Using the path provided by the previous iteration will significantly reduce computational burden as we have a good initial estimate of path than compared to beginning with  $\hat{v}_c^0(t) = [\hat{v}_0,\hat{v}_0 \cdots \hat{v}_0]$ and $\hat{\ddot{\theta}}^0(t) = [\hat{\ddot{\theta}}_0, \hat{\ddot{\theta}}_0 \cdots \hat{\ddot{\theta}}_0]$.

%\begin{itemize}
%\item At the beginning of optimization i.e., when $k=0$ we start with $\hat{v}_c^0(t) = [\hat{v}_0,\hat{v}_0 \cdots \hat{v}_0]$ and $\hat{\ddot{\theta}}^0(t) = [\hat{\ddot{\theta}}_0, \hat{\ddot{\theta}}_0 \cdots \hat{\ddot{\theta}}_0]$ and a $x_f, y_f$ provided by the high level planner such as \cite{sharmas}.  
%
%\item Once we have executed first $m$ control outputs provided by the optimization routine we observe the current state, obstacle information through on-board sensors and use this output as an initial estimate for our next iteration.
%
%\item Using the path provided by the previous iteration will significantly reduce computational burden as we have a good initial estimate of path than compared to beginning with  $\hat{v}_c^0(t) = [\hat{v}_0,\hat{v}_0 \cdots \hat{v}_0]$ and $\hat{\ddot{\theta}}^0(t) = [\hat{\ddot{\theta}}_0, \hat{\ddot{\theta}}_0 \cdots \hat{\ddot{\theta}}_0]$.
%
%\end{itemize}

\begin{algorithm}[!tbh]
 \caption{Alternating Optimization}\label{algo1}
    \begin{algorithmic}[1]
    \State  \textbf{Initialization}: Initial guess for $\hat{v}_c^k(t)$, $\hat{\ddot{\theta}}^k(t)$, iteration counter, $k=0$, $w_{\theta}, w_{v}$ \\
    
    \State $(\hat{x}^k(t), \hat{y}^k(t))=InitialTraj(\hat{v}_c^k(t)$, $\hat {\ddot{\theta}}^k(t))$\\

      \While {$\vert J^{k+1}_{\theta}-J^{k}_{\theta}\vert \geq \epsilon$ and $\vert J^{k+1}_{v}-J^{k}_{v}\vert \geq \epsilon$}      
    \State 
\begin{subequations}
\begin{align}
\ddot{\theta}^k(t) = \arg\min J_{\theta} + \sum w_{\theta}s_{\theta}.\nonumber \\
\textbf{x}(t_{i+1}) = \textbf{f}^{\theta}(\textbf{x}(t_i), \textbf{u}(t_i)). \nonumber \\
\dot{\theta}_{min} \leq \dot{\theta}(t_i)\leq \dot{\theta}_{max}. \nonumber \\
\ddot{\theta}_{min}\leq \ddot{\theta}(t_i)\leq \ddot{\theta}_{max}\nonumber\\
-\kappa_{max}\hat{v}_c(t_i)\leq \dot{\theta}(t_i)\leq \kappa_{max}\hat{v}_c(t_i).\nonumber \\
^{affine}c_{obst}(\ddot{\theta}(t_i))-s_{\theta}(t_i)\leq 0.\nonumber\\
s_{\theta}(t_i)\geq 0. \nonumber\\
{\theta}^{k-1}(t_i)-{\theta}^{trust}(t_i)\leq {\theta}^k(t_i) \leq {\theta}^{k-1}(t_i) + {\theta}^{trust}(t_i).\nonumber  
\end{align}
\end{subequations}
\If{$c_{obst}(.)>0$}
\State $w_{\theta} \leftarrow w_{\theta}*\delta$
\EndIf \\
\State $\hat{\ddot{\theta}}^{k}(t)\leftarrow \ddot{\theta}^k(t)$
\State 
\begin{subequations}
\begin{align}
v_c^k(t) = \arg\min J_{v} +\sum w_{v}s_{v}.\nonumber \\
\textbf{x}(t_{i+1}) = \textbf{f}^{v}(\textbf{f}(t_i), \textbf{u}(t_i)). \nonumber \\
v_c(t_i)\leq v_{max}. \nonumber\\
a_{min} \leq \frac{v_c(t_{i+1})-v_c(t_{i})}{\Delta t} \leq a_{max} \nonumber\\
-\kappa_{max}v(t_i)\leq \hat{\dot{\theta}}^k(t_i)\leq \kappa_{max}v(t_i).\nonumber \\
^{affine}c_{obst}(v_c(t_i))-s_{v}(t_i)\leq 0.\nonumber\\
s_{v}(t_i)\geq 0. \nonumber
\end{align}
\end{subequations}
\If{$c_{obst}(.)>0$}
\State $w_{v} \leftarrow w_{v}*\delta$
\EndIf \\   
\State $\hat{v}_c^{k}(t)\leftarrow v_c^k(t)$
\State $update \ \tau(\hat{v}^{k}(t))$    
    \State $k \leftarrow k+1$
    \EndWhile
        \end{algorithmic}  
        \normalsize 
        \end{algorithm}
  
\section{Results and Discussion}
\label{results_disc}

\subsection{Benchmarking Alternating Minimization (AM)}
\begin{figure}[!tbh]
  \centering
   \subfigure[]{
    \includegraphics[width= 4.25cm, height=3.5cm] {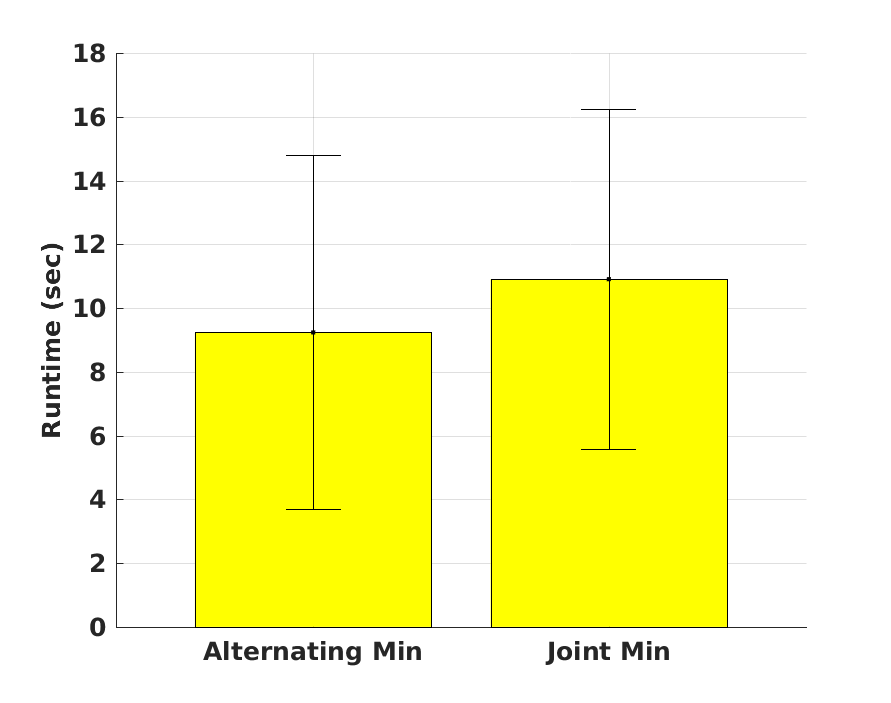}
    \label{opt_1}
   }\hspace{-0.5cm}
   \subfigure[]{
    \includegraphics[width= 4.25cm, height=3.5cm] {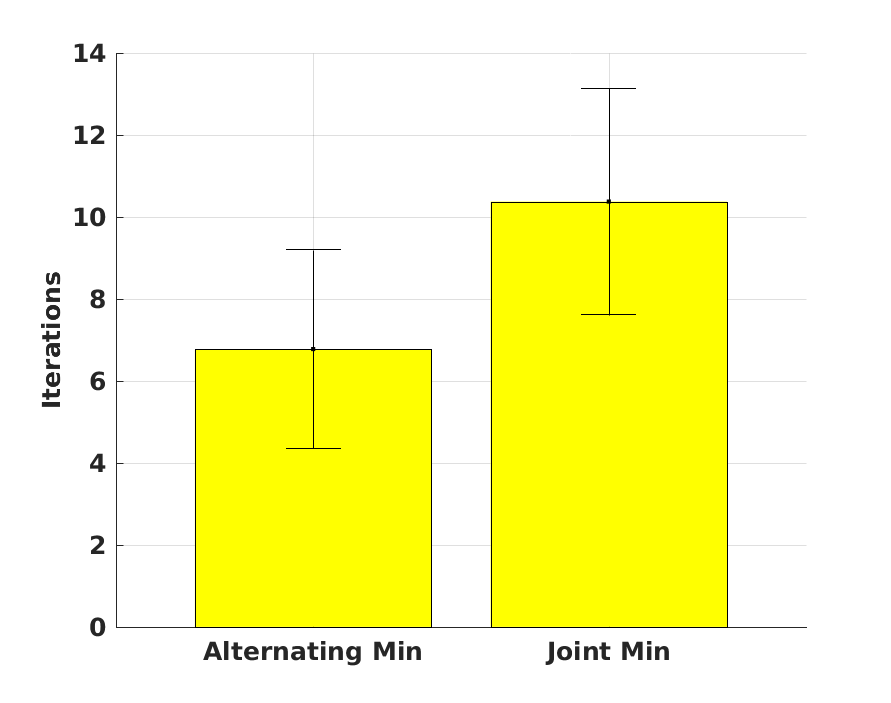}
    \label{opt_2}
   }
   \caption{$(a)$ shows a comparison of runtime of both approaches. Proposed approach shows $1.8 sec$ improvement in runtime. $(b)$ shows comparison on number of iterations taken by both approaches in-order to converge to a similar solution -compared in \ref{opt_33}-\ref{opt_34}}          
\end{figure}

\begin{figure}[!tbh]
  \centering
   \subfigure[]{
    \includegraphics[width= 4.25cm, height=3.5cm] {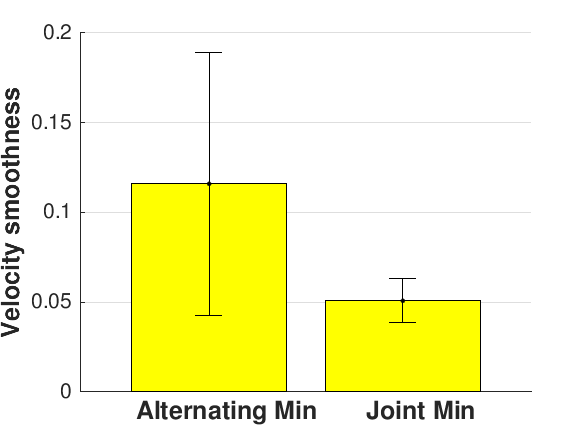}
    \label{opt_33}
   }\hspace{-0.5cm}
   \subfigure[]{
    \includegraphics[width= 4.25cm, height=3.5cm] {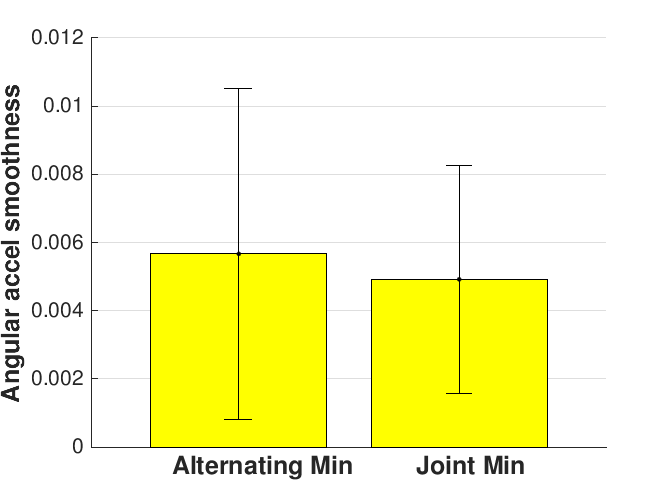}
    \label{opt_34}
   }
   \caption{$(a)$ and $(b)$ shows comparison of velocity smoothness and angular acceleration smoothness respectively, We observe reduction in smoothness of control profile due to alternating minimization}          
\end{figure}

\noindent Here we compare our proposed AM with the more conventional formulation where angular accelerations and linear velocities are simultaneously obtained. We prototyped both the approaches in MATLAB using CVX \cite{cvx_matlab}. These simulations are ran in a Intel core i5-3230M @ 2.6 GHz
CPU with 6GB RAM. The results are summarized in Fig.\ref{opt_1}-\ref{opt_2} and \ref{opt_33}-\ref{opt_34}.

\begin{figure*}[!tbh]
  \centering
   \subfigure[]{
    \includegraphics[width= 3.6cm, height=1.8cm] {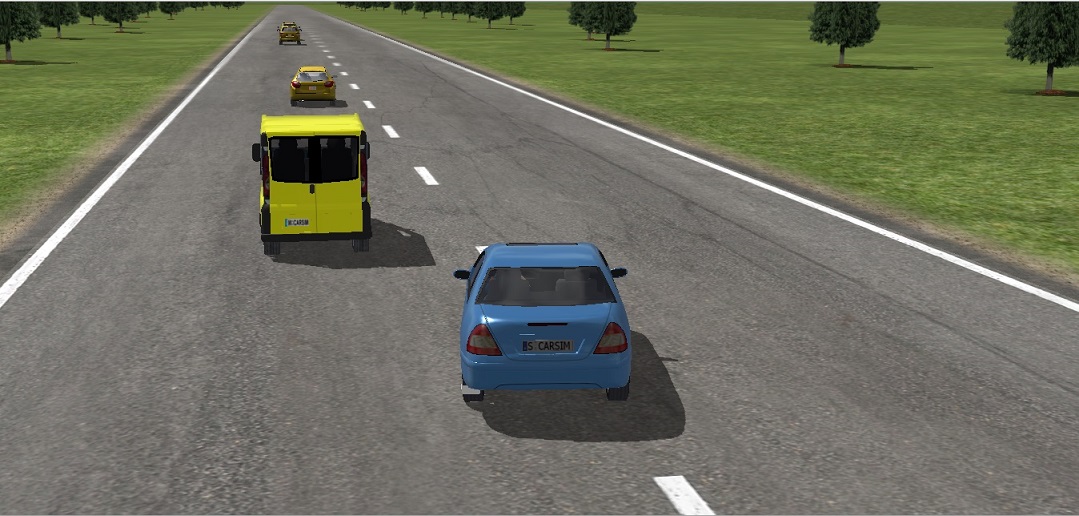}
    \label{occ_1}
   }
   \subfigure[]{
    \includegraphics[width= 3.6cm, height=1.8cm] {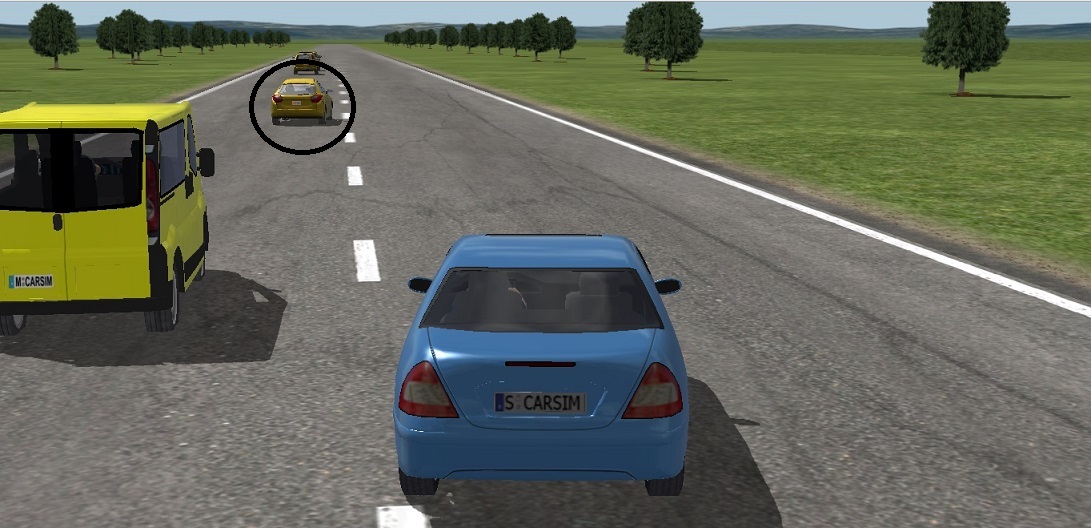}
    \label{occ_2}
   }
    \subfigure[]{
    \includegraphics[width= 3.6cm, height=1.8cm] {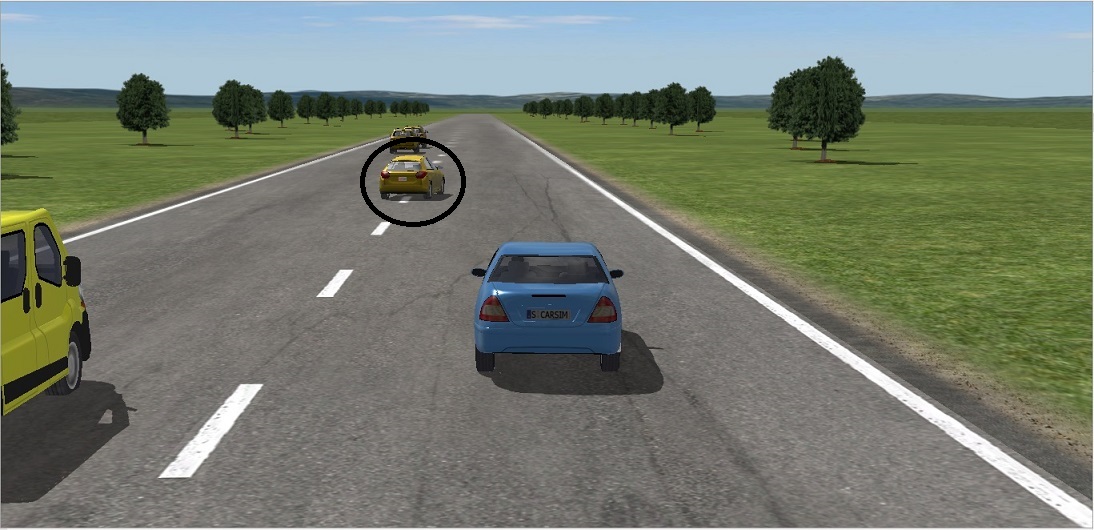}
    \label{occ_3}
   }
    \subfigure[]{
    \includegraphics[width= 3.6cm, height=1.8cm] {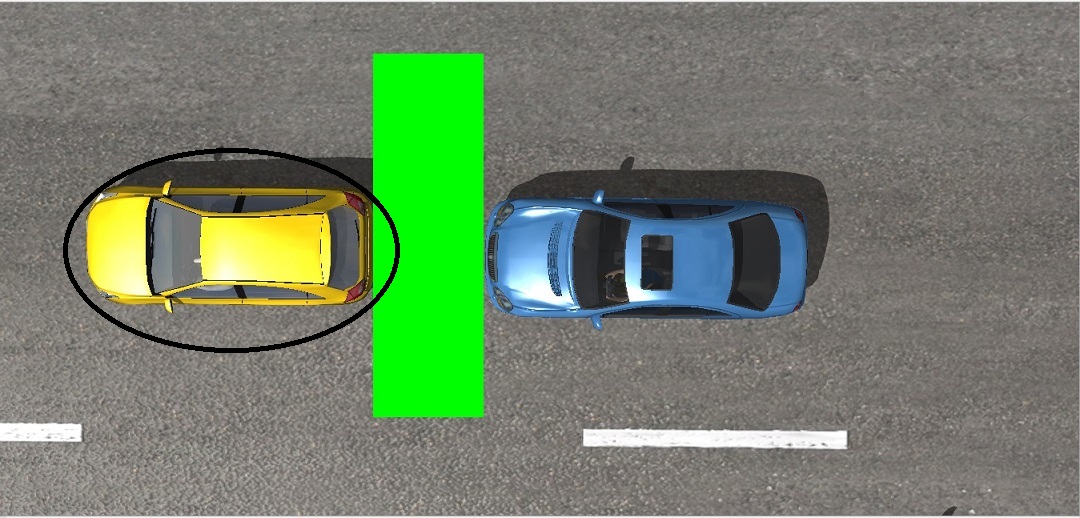}
    \label{occ_4}
   }
   \subfigure[]{
    \includegraphics[width= 3.6cm, height=1.8cm] {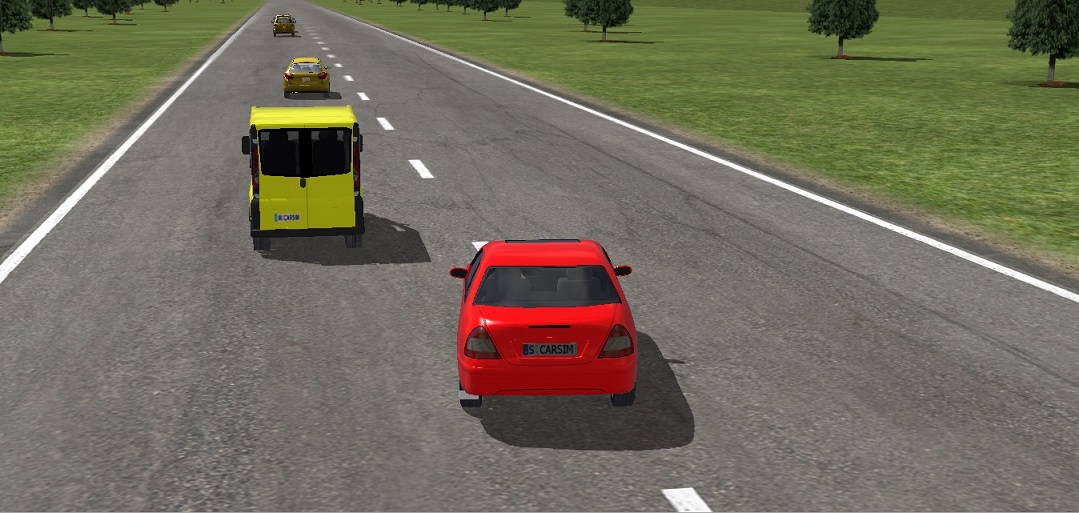}
    \label{occ_5}
   }
   \subfigure[]{
    \includegraphics[width= 3.6cm, height=1.8cm] {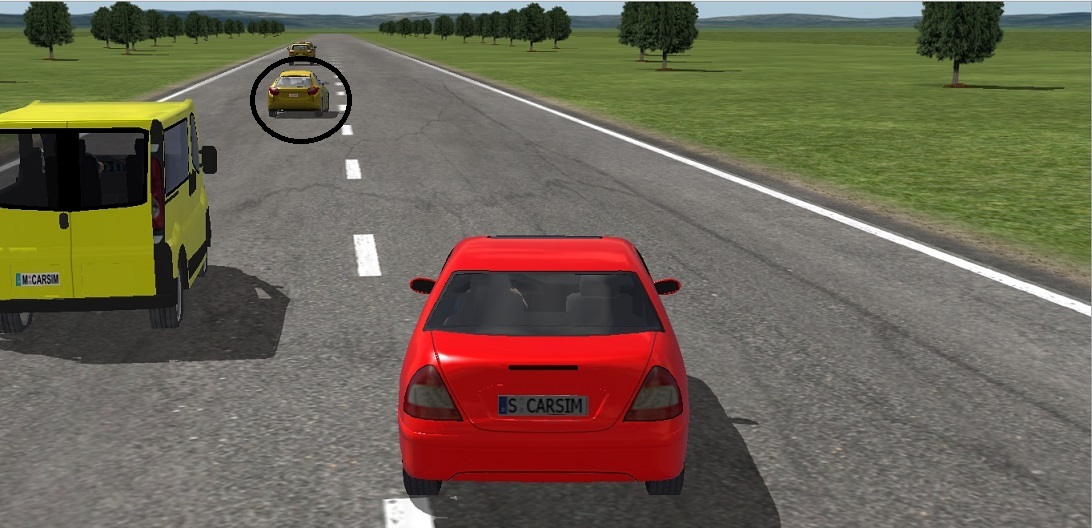}
    \label{occ_6}
   }
    \subfigure[]{
    \includegraphics[width= 3.6cm, height=1.8cm] {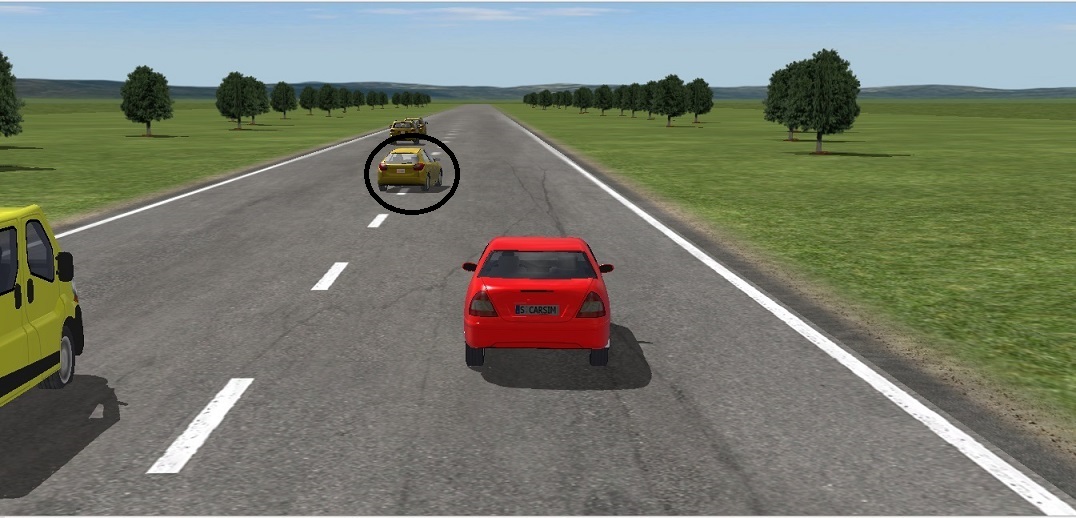}
    \label{occ_7}
   }
    \subfigure[]{
    \includegraphics[width= 3.6cm, height=1.8cm] {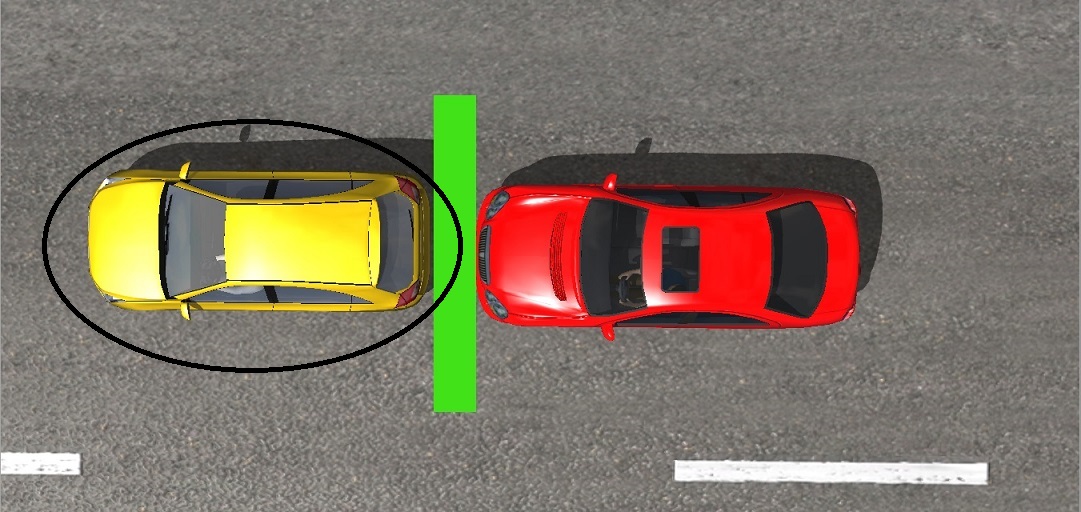}
    \label{occ_8}
   }
   \caption{ (a)-(d) and (e)-(f) shows a scenario where in an occluded vehicle (marked with black circle) makes a sudden overtaking maneuver and comes within collision range of an ego vehicle. Vehicle in blue is modelled with actuator dynamics as in eqn. (\ref{first_order_model}) and Vehicle in red is modelled with actuator dynamics as in eqn. (\ref{linear_act}). By comparing (d) and (h) we observe a significant improvement in inter-vehicle distance(marked in green)}          
\end{figure*}

Fig.\ref{occ_1}-\ref{occ_2} presents the comparison of the computational aspects in terms of run time and number of iterations. As shown, on an average, our proposed AM approach takes around $17 \%$ less time and $32 \%$ less iterations. As shown in Fig.\ref{opt_33}-\ref{opt_34}, both our AM and joint formulation results in low smoothness cost. However, smoothness cost obtained with joint formulation is significantly lower than that obtained with our AM. 
It can be thus concluded that our proposed AM approach sacrifices a bit of optimality in the pursuit of computing reasonably smooth, kinematically feasible collision avoiding trajectories in quick time. We believe such a feature can be useful for autonomous driving.

%We follow the way-points placed along the lanes of road to simulate different driving speeds along these lanes. 

\subsection{Effect of Actuator Dynamics}
\noindent To analyze the effect of actuator dynamics, we integrated our MPC with a state of the art vehicle simulator called CARSIM. The simulator provides state feedback for the ego-vehicle along with the information about the state of the obstacle through a virtual LIDAR with a sensing range of $70m$. For this implementation, we prototyped Algorithm \ref{algo1} in C++ using Gurobi solver \cite{gurobi} obtaining significant speed up. We were able to iterate our MPC at 10Hz with a planning horizon of 50 steps each of duration 0.1s. To ensure that the ego vehicle respects the road geometry, the boundaries of the road are modeled as imaginary obstacles and included into Algorithm \ref{algo1}. The bounds on velocity ($v_{max}$), acceleration($a_{max}$), and deceleration($a_{min}$) were kept at $25m/s$, $4 m/s^{2}$, and $-6 m/s^{2}$ respectively to make simulations closer to real-world scenarios. The detailed videos of simulations in several safety critical situations described in Section \ref{benchmark} are provided at {\textbf{\url{https://researchweb.iiit.ac.in/~mithun.babu/acc_2019.mp4}}}

%improves inter vehicle distance from $0.6 m$ to $1.52 m$ as can be seen from Fig. \ref{occ_4}, \ref{occ_8} and \ref{occ_sd}. 

\begin{figure}[!h]
  \centering
   \subfigure[]{
    \includegraphics[width= 3in, height=1in] {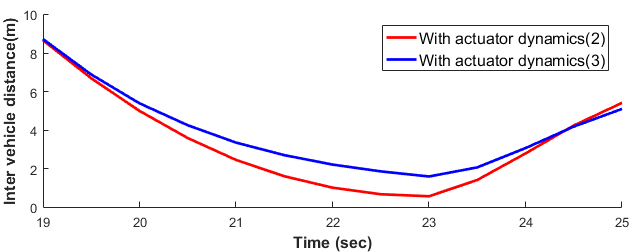}
    \label{occ_sd}
   }
   \subfigure[]{
    \includegraphics[width= 3in, height=1in] {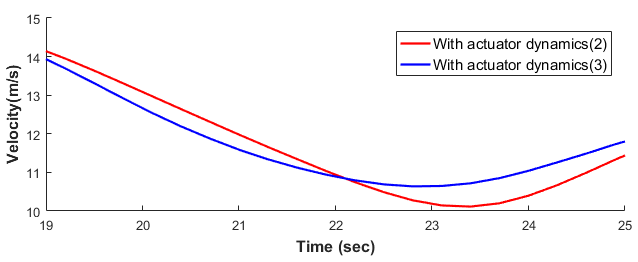}
    \label{occ_vel_zoom}
   }
   \subfigure[]{
    \includegraphics[width= 3in, height=1in] {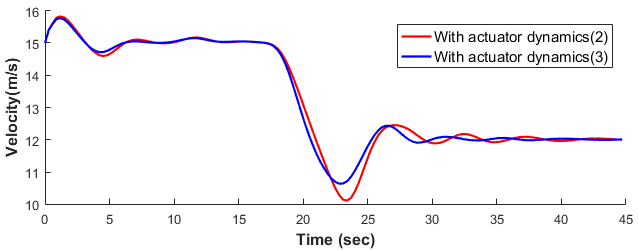}
    \label{occ_vel}
   }
   \caption{ (a) shows inter vehicle distance between ego vehicle and the leading vehicle during critical situation using both actuator models. (b) and (c) show the velocity plot during the maneuver. We can notice advantage of our approach in terms of decreased velocity overshoot and improved damping.}          
\end{figure}

\begin{figure*}[tbh]
  \centering
\subfigure[]{
    \includegraphics[width= 3.6cm, height=1.8cm] {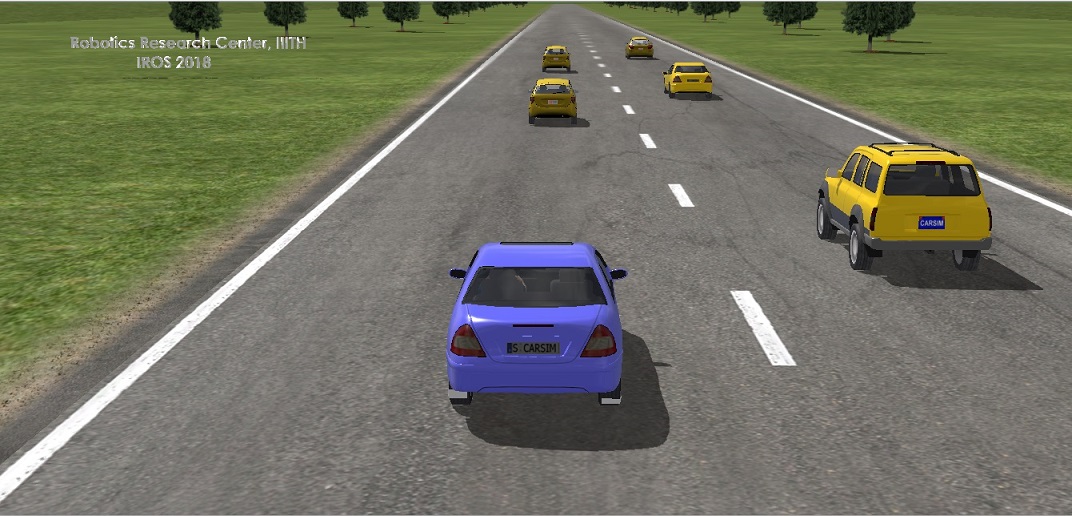}
    \label{lcc_1}
   }
   \subfigure[]{
    \includegraphics[width= 3.6cm, height=1.8cm] {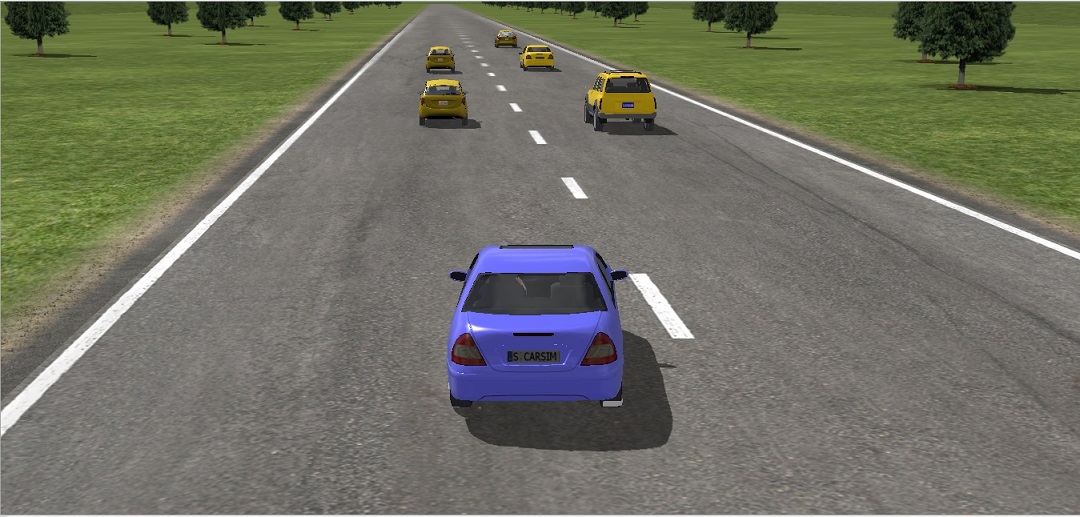}
    \label{lcc_2}
   }
    \subfigure[]{
    \includegraphics[width= 3.6cm, height=1.8cm] {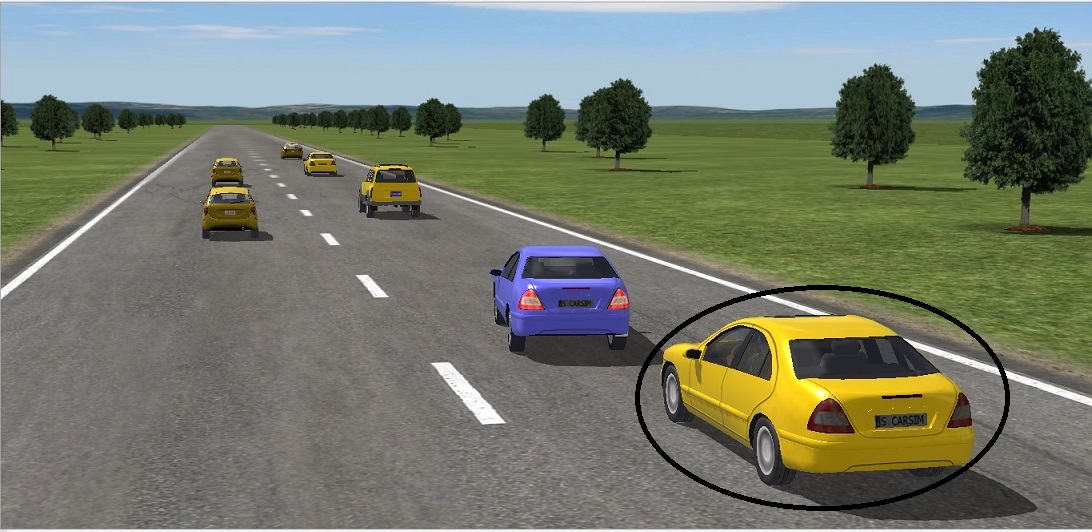}
    \label{lcc_3}
   }
    \subfigure[]{
    \includegraphics[width= 3.6cm, height=1.8cm] {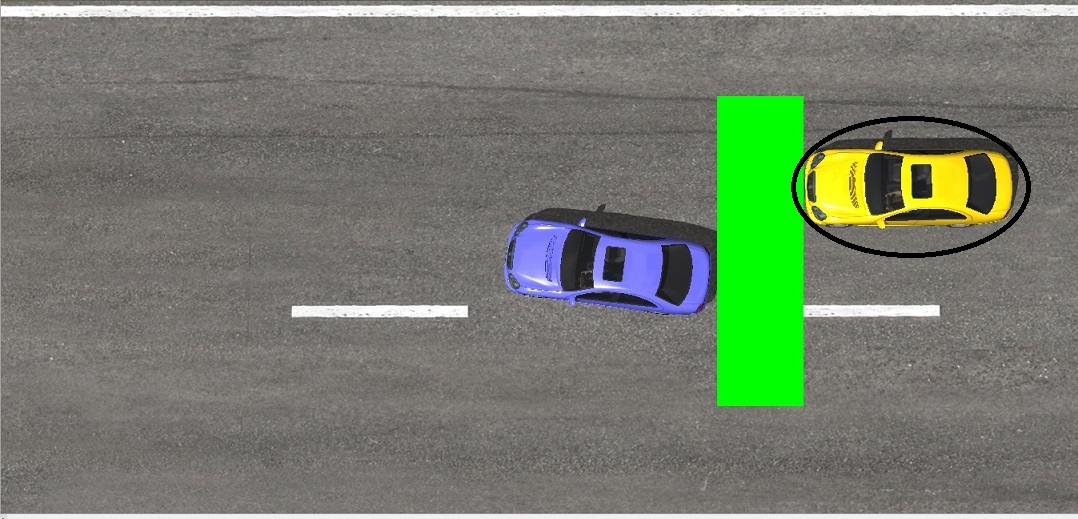}
    \label{lcc_4}
   }   
   \subfigure[]{
    \includegraphics[width= 3.6cm, height=1.8cm] {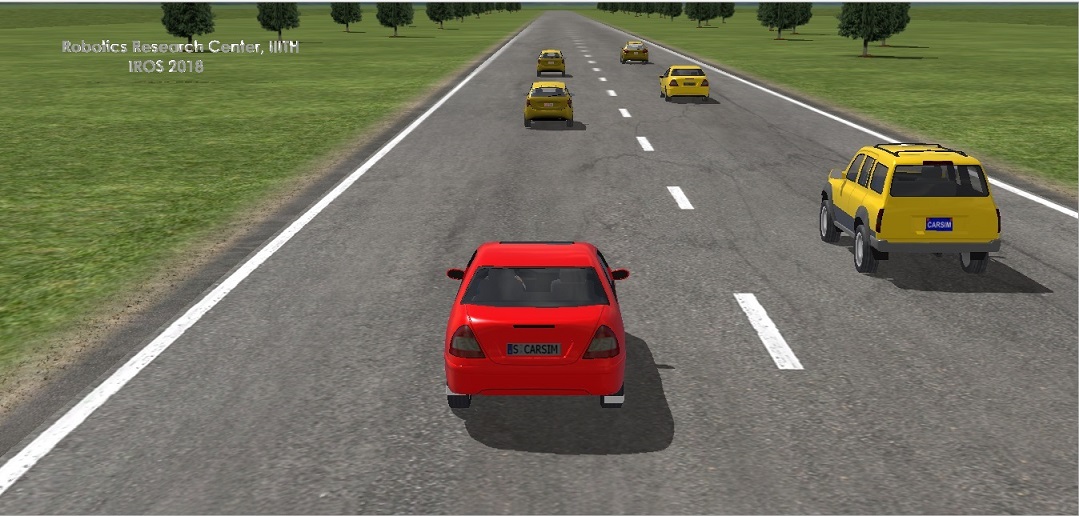}
    \label{lcc_5}
   }
   \subfigure[]{
    \includegraphics[width= 3.6cm, height=1.8cm] {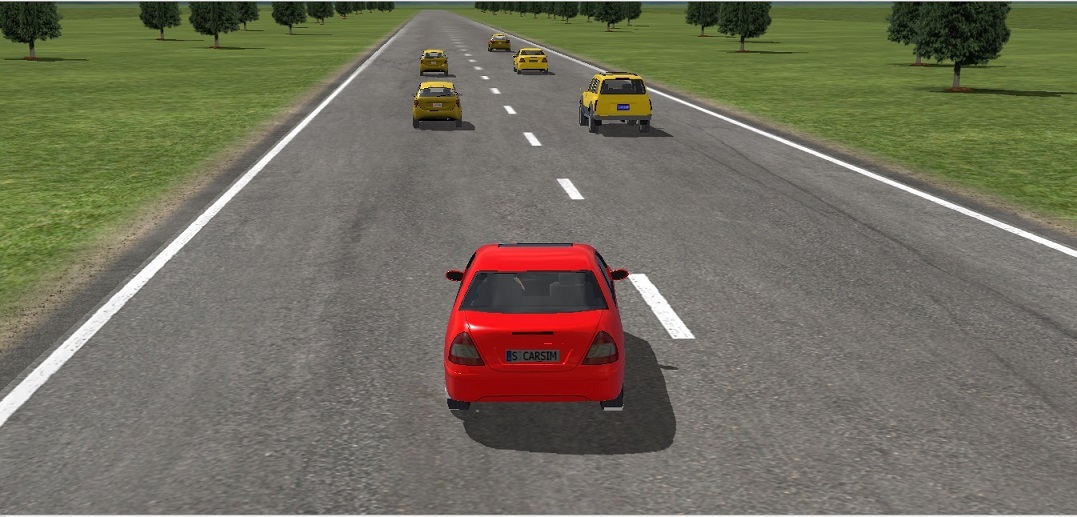}
    \label{lcc_6}
   }
    \subfigure[]{
    \includegraphics[width= 3.6cm, height=1.8cm] {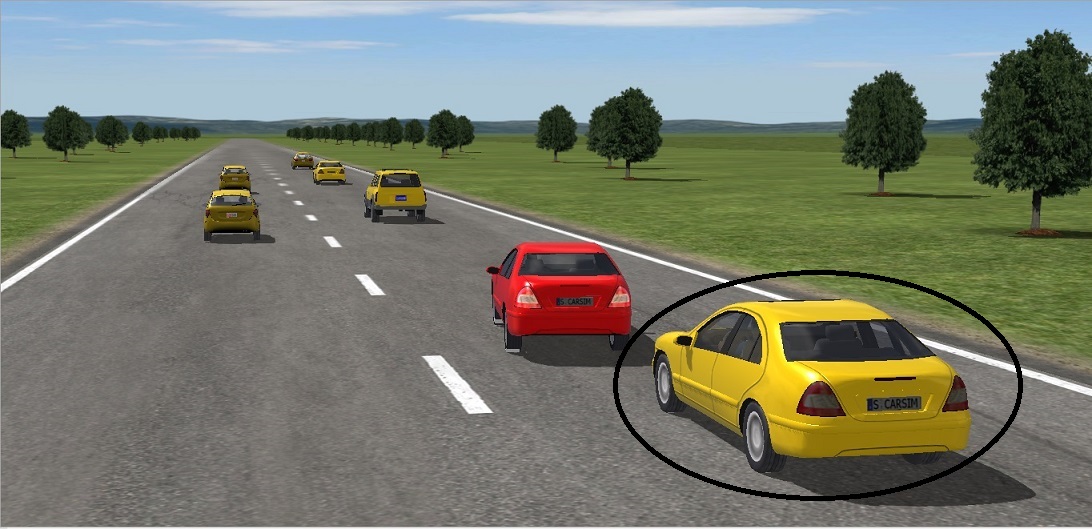}
    \label{lcc_7}
   }
    \subfigure[]{
    \includegraphics[width= 3.6cm, height=1.8cm] {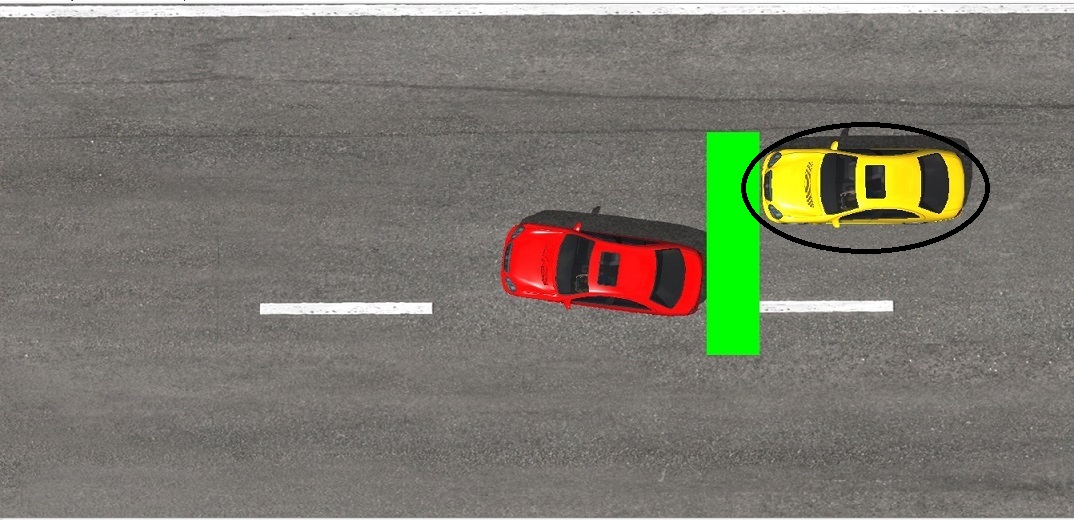}
    \label{lcc_8}
   }   
   \caption{ (a)-(d)  shows a lane change scenario with an ego vehicle (blue) with actuator dynamics modelled as in eqn.(\ref{first_order_model}). While the ego vehicle is executing the lane change, a vehicle from the rear (marked with black circle) suddenly comes into the collision range of the ego vehicle. The shaded region in green is an indicator of the inter-vehicle distance.  (e)-(f) repeats the same experiment with an ego vehicle (red) with actuator dynamics modelled as in eqn.(\ref{linear_act}). The improvement in the inter-vehicle distance can be seen in (d).}          
\end{figure*}

\begin{figure}[!h]
  \centering
   \subfigure[]{
    \includegraphics[width= 3in, height=1in] {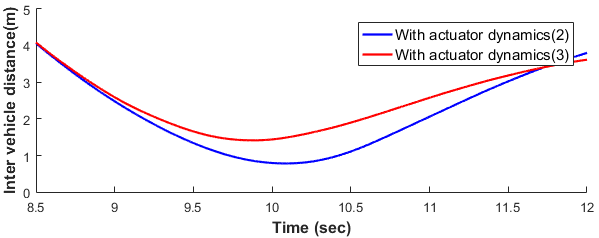}
    \label{lcc_sd}
   }
   \subfigure[]{
    \includegraphics[width= 3in, height=1in] {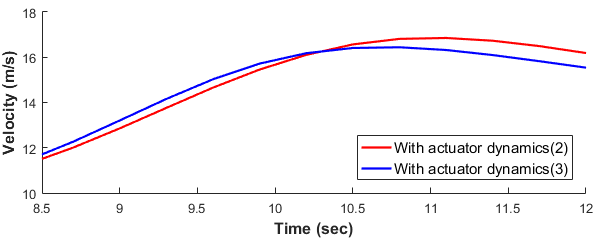}
    \label{lcc_vel_zoom}
   }
   \subfigure[]{
    \includegraphics[width= 3in, height=1in] {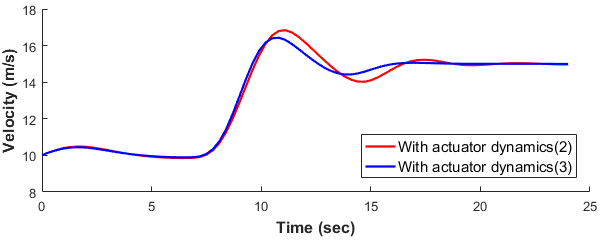}
    \label{lcc_vel}
   }
   \caption{(a) plots inter vehicle distances between the ego vehicle and obstacle in rear using both actuator models. The blue curve maintains better inter vehicle distances compared to the red curve, thus validating our approach. (c) signifies the advantage of our approach in terms of decreased velocity overshoot and improved damping. }          
\end{figure}

\begin{figure}[!tbh]
  \centering
    \subfigure[]{
    \includegraphics[width= 3.6cm, height=1.8cm] {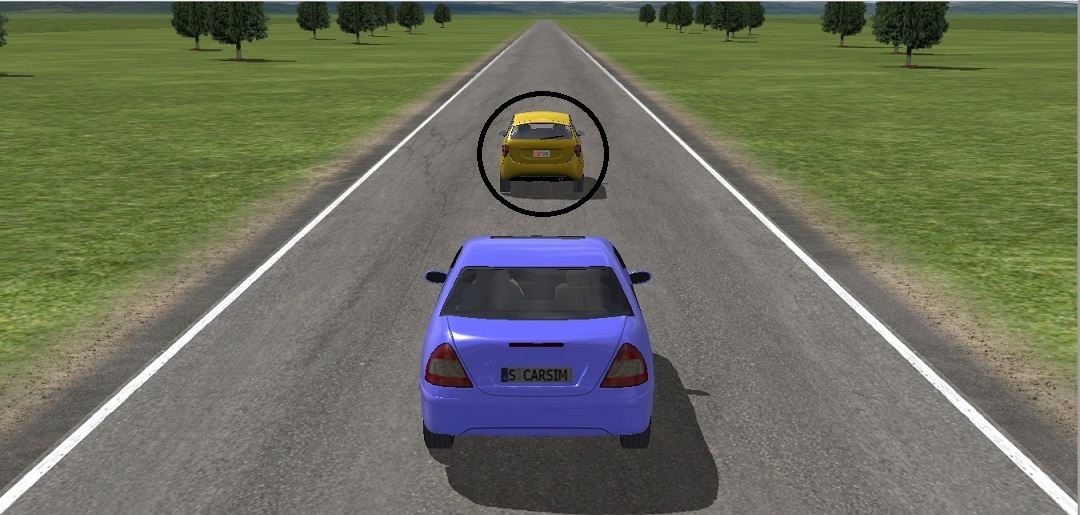}
    \label{sbr_b_1}
   }
   \subfigure[]{
    \includegraphics[width= 3.6cm, height=1.8cm] {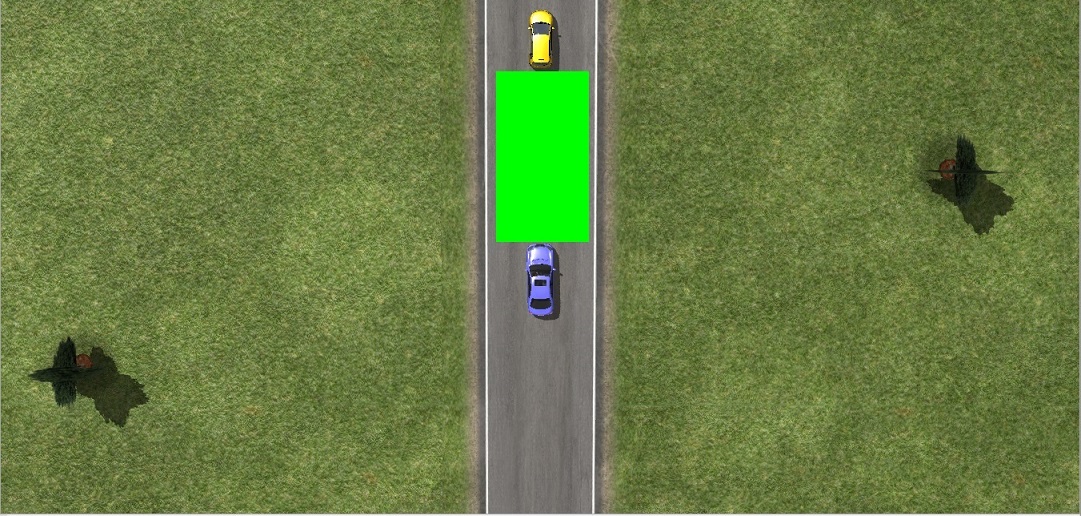}
    \label{sbr_b_2}
   }  
   \subfigure[]{
    \includegraphics[width= 3.6cm, height=1.8cm] {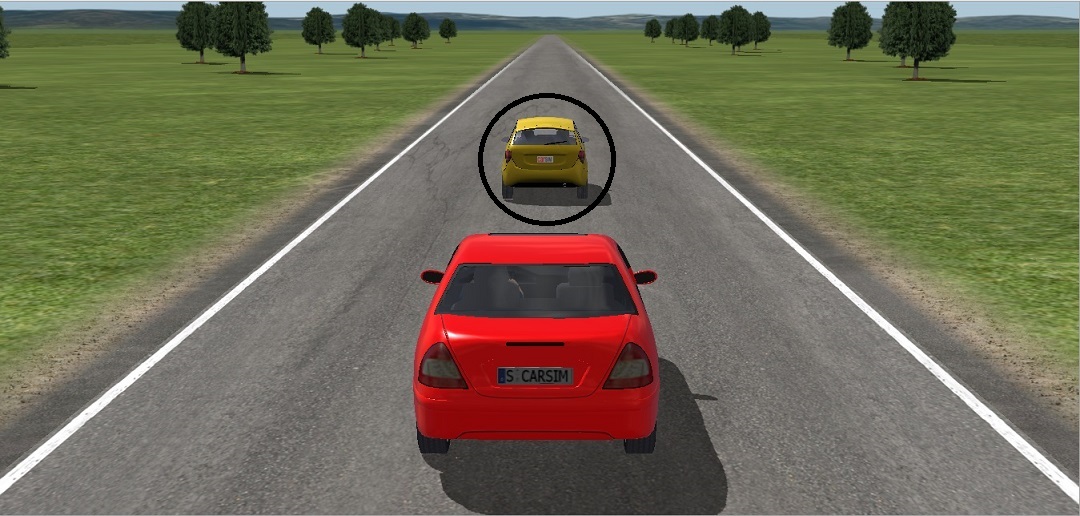}
    \label{sbr_1}
   }
   \subfigure[]{
    \includegraphics[width= 3.6cm, height=1.8cm] {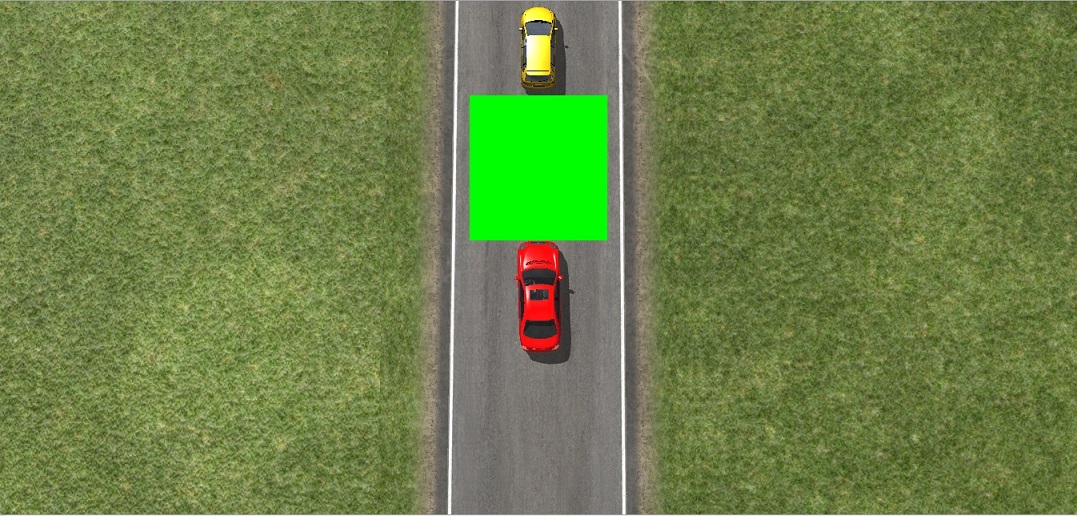}
    \label{sbr_2}
   }
   \caption{ (a)-(b) show a scenario where an ego vehicle (blue) tries to follow the yellow vehicle (marked in black circle) ahead. Later, this vehicle brake abruptly brakes and consequently comes into the collision range. In (c)-(d), we show the performance of the ego vehicle(red) with actuator dynamics modelled as eqn.(\ref{linear_act}). The improvement in the inter-vehicle distance is shown by comparing (b) with (d).}          
\end{figure}

\begin{figure}[!tbh]
  \centering
   \subfigure[]{
    \includegraphics[width= 3in, height=1in] {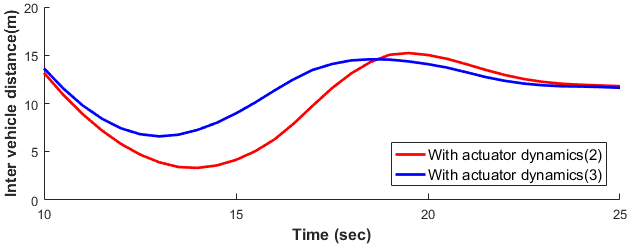}
    \label{sbr_sd}
   }
   \subfigure[]{
    \includegraphics[width= 3in, height=1in] {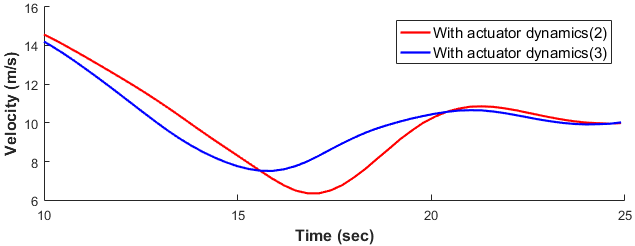}
    \label{sbr_vel_zoom}
   }
   \subfigure[]{
    \includegraphics[width= 3in, height=1in] {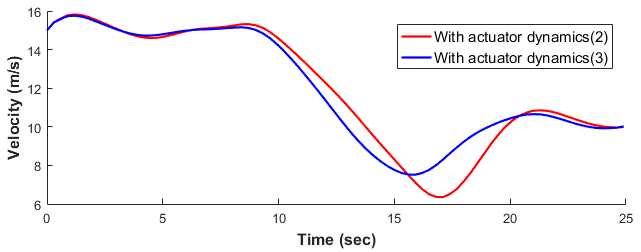}
    \label{sbr_vel}
   }
   \caption{(a) plots inter vehicle distance between the ego vehicle and leading vehicle using two different models of actuator dynamics. (c) further shows advantages of our approach in terms of decreased velocity overshoot and improved damping.}          
\end{figure}

\begin{figure}[!h]
\includegraphics[width= 3in, height=1.5in] {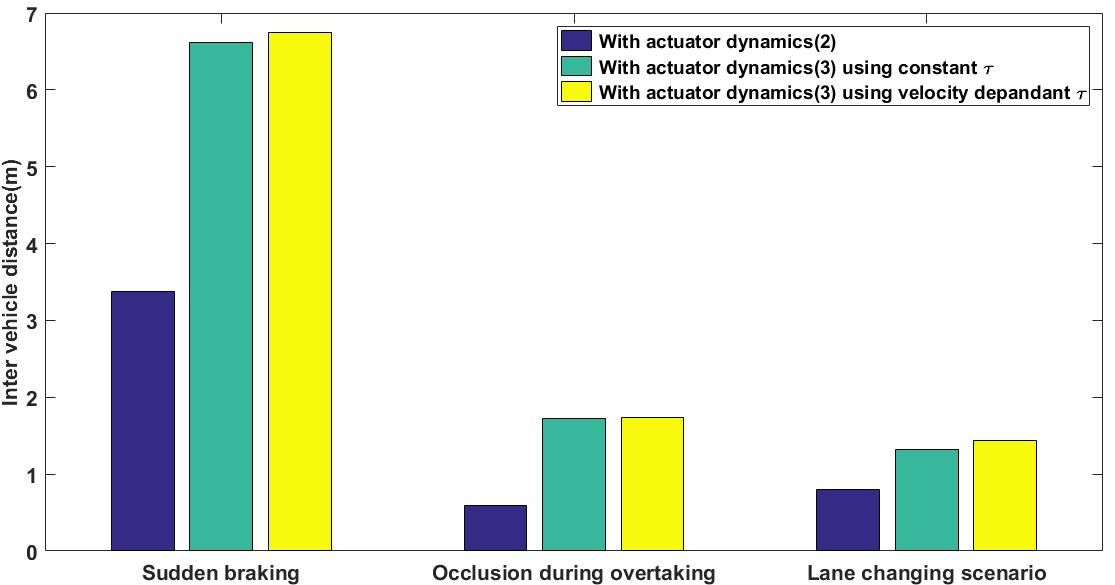}
   \caption{Figure summarizes results on inter-vehicle distance obtained in previous benchmarks. It is clear that the our MPC based on AM and actuator dynamics (\ref{first_order_model}) has a clear advantage while maneuvering in typical urban scenarios}
   \label{bar_safe}
\end{figure}

\subsubsection{Benchmark Scenarios}\label{benchmark}
\noindent \textbf{Occlusion during overtaking:} In this benchmark, the ego-vehicle traveling at $15m/s$ initiates an overtaking maneuver. Midway during the maneuver, it notices another vehicle (which was previously occluded from its field of view) also performing a similar maneuver. In such a situation, the ego-vehicle should decelerate quickly to $12 m/s$ to avoid collision. Fig. \ref{occ_1}-\ref{occ_4} shows the occlusion scenario, wherein the ego-vehicle is shown in blue. The vehicle it is overtaking is the yellow van and the occluded vehicle is marked with a black circle. Fig.\ref{occ_5}-\ref{occ_8} repeats the same benchmark with an ego-vehicle (shown in red) that uses MPC with actuator dynamics (\ref{linear_act}) to compute its motion. Fig.\ref{occ_sd},\ref{occ_vel_zoom} shows the plot of inter-vehicle distance and vehicle velocity for a specific portion of time. It can be seen that the ego-vehicle that incorporates actuator dynamics (\ref{first_order_model}) within MPC can pro-actively anticipate the transient velocity response and thus de-accelerate faster resulting in improved inter-vehicle distance. The complete velocity profile is shown in Fig.\ref{occ_vel} and as shown it exhibits less overshoot and oscillation with the incorporation of the actuator dynamics.

\noindent \textbf{Lane change :} In this benchmark, the ego-vehicle moving at a speed of $10m/s$ performs a lane change maneuver to an adjacent high speed lane ($15m/s$). In this situation, a delay in achieving the increased velocity would bring the ego-vehicle in collision course with the vehicle approaching from the rear. Fig.\ref{lcc_1}-\ref{lcc_4} show the lane change scenario. The ego-vehicle is shown in blue while the rear vehicle is marked with a black circle. Similar to the previous benchmark, in Fig.\ref{lcc_5}-\ref{lcc_8}, we repeat the simulation with a different ego-vehicle (shown in red) that has actuator dynamics modeled as eqn.(\ref{linear_act}). It can be seen from Fig.\ref{lcc_vel_zoom} that use of actuator dynamics (\ref{first_order_model})  allows the vehicle to accelerate faster to the required velocity and thus maintain higher clearance from the vehicle approaching from behind (Fig.\ref{lcc_sd}). The complete vehicle velocity shown in Fig.\ref{lcc_vel} shows improved transient response with reduced overshoot and oscillation when using actuator dynamics (\ref{first_order_model}) within the MPC.

\noindent \textbf{Sudden Braking:}
This benchmark is shown in Fig.\ref{sbr_b_1}-\ref{sbr_b_2}. Here, the ego-vehicle (shown in blue) is following a yellow vehicle (marked by a black circle) in the front at a speed of $15 m/s$. Suddenly, the front vehicle de-accelerates bringing our ego-vehicle in collision course. The ego-vehicle responds by reducing its speed. Fig.\ref{sbr_1}-\ref{sbr_2} repeats the same simulation with an ego-vehicle (shown in red) with actuator dynamics  (\ref{linear_act}). As shown in Fig.\ref{sbr_vel_zoom}, bringing actuator dynamics explicitly within the MPC allowed the ego-vehicle to achieve significantly faster de-acceleration. This in turn improves the inter-vehicle distance (Fig.\ref{sbr_sd}). Similar to previous benchmarks, in Fig.\ref{sbr_vel}, we observe velocity profile with improved transient response with actuator dynamics (\ref{first_order_model}) incorporated within the MPC.

\noindent \textbf{Quantitative Comparison:} The bar-graph shown in Fig. \ref{bar_safe} summarizes the results on inter vehicle distance observed during previous benchmarks. Here, we also present an additional comparison that is based on actuator dynamics (eqn.\ref{first_order_model}) but with a $\tau$ which varies with vehicle velocity $v(t)$. We used a function approximator based on radial basis function to to learn the $\tau-v(t)$ mapping. As shownn in Fig.\ref{bar_safe}, we noticed only marginal benefit of using a complicated $v(t)$ dependent $\tau$ in our first order actuator dynamics (\ref{first_order_model}).

\section{Conclusions and Future Work}

In this work, we built an MPC formulation, on a novel alternating minimization based optimization routine coupled with a non-holonomic motion model with a first order actuator dynamics. We performed extensive simulations to show how incorporation of actuator dynamics within the MPC improves autonomous driving. In particular, we showed improved inter-vehicle distance during different maneuvers like overtaking, lane-change and vehicle-following. We also benchmarked our alternating minimization and showed that it can compute feasible collision avoiding trajectories faster while still achieving quite low smoothness cost.

There are several directions where our current formulation can be extended to remove some of the existing limitations. For example, one of our primary future focus is on incorporating steering actuator dynamics within our MPC and analyzing the resulting benefits in the context of autonomous driving. We are also working on implementing the proposed formulation on our autonomous car prototype.

\bibliographystyle{IEEEtran}  
\bibliography{ref_iros18}

\end{document}